# An adaptive bi-objective optimization algorithm for the satellite image data downlink scheduling problem considering request split

Zhongxiang Chang[1,2,3], Abraham P. Punnen[2*], Zhongbao Zhou[1,3]

[1]School of Business Administration, Hunan University, Changsha, China, 410082

[2]Department of Mathematics, Simon Fraser University, Surrey, BC, Canada, V3T 0A3

[3]Hunan Key Laboratory of intelligent decision-making technology for emergency management, Changsha, China, 410082

## Abstract

The satellite image data downlink scheduling problem (SIDSP) is well studied in literature for traditional satellites. With recent developments in satellite technology, SIDSP for modern satellites became more complicated, adding new dimensions of complexities and additional opportunities for the effective use of the satellite. In this paper, we introduce the dynamic two-phase satellite image data downlink scheduling problem (D-SIDSP) which combines two interlinked operations of image data segmentation and image data downlink, in a dynamic way, and thereby offering additional modelling flexibility and renewed capabilities. D-SIDSP is formulated as a bi-objective problem of optimizing the image data transmission rate and the service-balance degree. Harnessing the power of an adaptive large neighborhood search algorithm (ALNS) with a nondominated sorting genetic algorithm II (NSGA-II), an adaptive bi-objective memetic algorithm, ALNS+NSGA-II, is developed to solve D-SIDSP. Results of extensive computational experiments carried out using benchmark instances are also presented. Our experimental results disclose that the algorithm ALNS+NSGA-II is a viable alternative to solve D-SIDSP more efficiently and demonstrates superior outcomes based on various performance metrics. The paper also offers new benchmark instances for D-SIDSP that can be used in future research works on the topic.

Keyword: Scheduling; Satellite image data downlink scheduling problem; Dynamic phases; Bi-

[*] corresponding author: Abraham P. Punnen, email: apunnen@sfu.ca

# 1. Introduction

Satellite image data downlink scheduling problem (SIDSP) plays a crucial role in the mission planning operation of earth observation satellites (EOSs) (Li et al., 2014). With the increase in the number of on-orbit EOSs (Wang et al., 2020) and modern technological developments (Chang et al., 2020, Lu et al., 2021, Chang et al., 2021b), the traditional meaning of SIDSP is now shifted, and the role of effective downlink scheduling in satellite mission planning becomes increasingly important.

With the development of commercial remote sensing satellites (CRSSs) (Sai et al., 2018, Jakhu and Pelton, 2014), the number of on-orbit EOSs increased significantly and so are their observational capabilities. In particular, the sensors installed on modern EOSs are of "ultra-high resolution" (Jones, 2018, Jawak and Luis, 2013) and the new generation EOSs have more flexibility and degrees of freedom (pitch, roll and yaw) offering active imaging capabilities (Chang et al., 2020, Chang et al., 2019). In summary, modern EOSs can acquire large amount of image data more easily because of more on-orbit EOSs, higher resolution sensors, and more flexibility in attitude maneuvering. However, the number of receiving resources (ground stations and relay satellites) has not increased much (Pei-jun et al., 2008). These conditions are bound to produce more conflicts and difficulties for downloading image data, and D-SIDSP addresses precisely this problem.

Many existing research works on the traditional SIDSP implicitly assume that image data cannot be segmented (Karapetyan et al., 2015) and have to be transmitted in "First observed, First downlink (FOFD)" order (Wang et al., 2011). Under these conditions, SIDSP is equivalent to the satellite range scheduling problem (SRSP) (Luo et al., 2017, Marinelli et al., 2011, She et al., 2019). But with the development of on-board file systems in EOS (Huang et al., 2018), an original image data can be split into several fragments, referred to as "Segment", and then transmitted without prioritizing the capturing order, referred to as "rearrange". Because of this additional dimension of complexity, SIDSP is no longer a simple downlink request permutation problem (DRPP) (Karapetyan et al., 2015). The modern SIDSP is more complicated than SRSP (Zufferey et al., 2008, Vazquez and Erwin, 2014, Barbulescu et al., 2004) and the traditional SIDSP (Wang et al., 2011, She et al., 2019).

SIDSP is a major bottleneck that limits the exploitation of the full power of EOSs by capturing more image data. In this paper, we focus on SIDSP designed for modern satellites to exploit the power of modern technological advancements. In fact, we redefine SIDSP taking into consideration the flexibility offered by "Segment" and "Rearrange", and view it as a dynamic two-phase satellite image data downlink scheduling problem (D-SIDSP) (See Figure 1).

The first phase of D-SIDSP is similar to the satellite range scheduling problem (SRSP), which is primarily used to optimize the transmission scheme and generates the downlink tasks for each EOS. The second phase is analogous to a one-dimensional two-stage cutting stock problem (TSCSP) (Muter and Sezer, 2018), where the original image data corresponds to 'stock rolls' in the cutting stock problem and the available transmission windows, (more precisely, the generated downlink tasks) correspond to 'finished rolls'. In addition, all stock rolls are first cut into intermediate rolls (segmented image data) whose widths are not known a priori but are restricted to lie within a specified interval, and then, in the second stage, finished rolls of demanded widths are produced from these intermediate rolls.

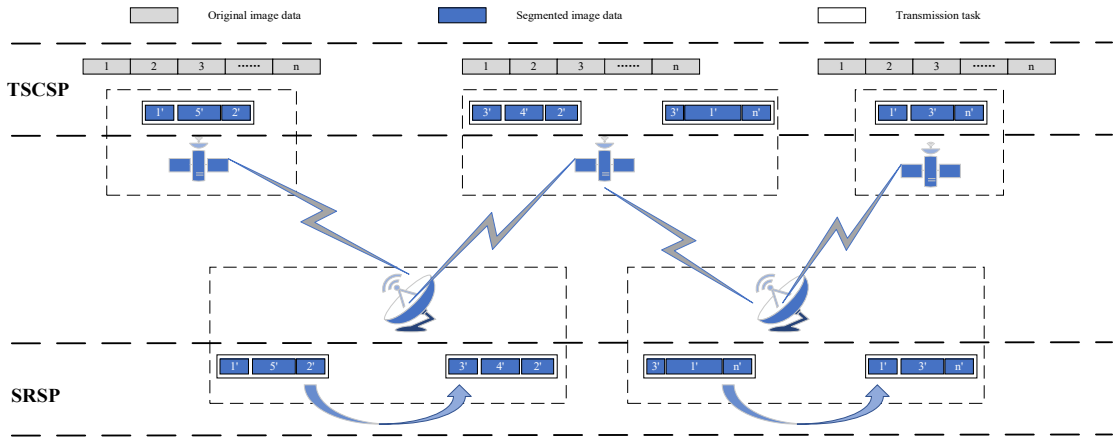

Figure 1 The sketch map for two phases of D-SIDSP

There are several existing research works and solution approaches for each phase of D-SIDSP when taken individually. However, the two phases of D-SIDSP are interrelated and influence the outcome of each other in a dynamic way. Therefore, D-SIDSP is a novel scheduling problem with two dynamic phases, and are more complicated than the traditional SIDSP, which itself is an NP-hard problem (Barbulescu et al., 2004, Vazquez and Erwin, 2014). Therefore, in this paper, we explore D-SIDSP in depth with the aim of transmitting increased quantity of image data and balance the service rate of all EOSs involved, exploiting the modern "Segment & rearrange" capabilities.

The rest of the paper is organized as follows. Section 2 presents a detailed description of D-SIDSP. In section 3, a bi-objective optimization model is proposed for D-SIDSP. Section 4 deals with the development of an adaptive bi-objective memetic algorithm, ALNS+NSGA-II, to solve the D-SIDSP. Section 5 reports on the results of extensive computational experiments carried out along with a detailed performance analysis of individual components of our algorithm. Concluding remarks are given in Section 6.

Throughout this paper, we represent variable and parameter names using one or more operators following the convention of the object-oriented programming framework which we believe will improve the readability and help ease of presentation of this complex problem. Note that these variables/parameters can alternatively be represented using one or more subscripts or superscripts. We preferred the object-oriented representation rather than the traditional mathematical representation of the constraints. For example, let $S$ be a parameter having attributes $x$ and $y$. Then, the x associated with $S$ is denoted by $S.x$ and y associated with $S$ is denoted by $S.y$. This notational convention is used throughout this paper without additional explanations.

## 2. Problem description and modeling

To better emphasize the primary focus in our paper, several reasonable assumptions are made to standardize and simplify D-SIDSP. These assumptions are consistent with existing research works in the field and current engineering practices. A closer examination of these assumptions discloses that they are not very restrictive, though appear as major restrictions in applicability at first sight.

(1) Though relay satellites can receive image data from EOSs, in reality, they are just agencies. In other words, image data received by relay satellites still need to be transmitted to ground stations. Therefore, we do not consider relay satellites as receiving resources in our study.

(2) We assume that there is only one antenna in each EOS and each ground station (GS). In

other words, a ground station can only receive image data from at most one EOS at any given time, and an EOS can only transmit image data to at most one GS at any given time.

(3) The original image data generated by an observation could be segmented into several fragments and then transmitted under different transmission windows, but all segmented image data should be transmitted completely, otherwise all of them should be abandoned. In other words, an original image data is considered in the downlink scheduling if and only if it can be transmitted completely during the planning horizon.

(4) As indicated earlier, since we have more observation capabilities than the transmission/receiving capabilities, we assume that an original image data can be transmitted at most once.

(5) The original image data needs to be transmitted from the EOS in a pre-specified and static way before we optimize the downlink scheme for D-SIDSP.

(6) Each EOS has sufficient on-board memory and enough power during the whole scheduling horizon of D-SIDSP and hence it is not necessary to work on memory management and acquiring power by dedicated charging operations.

(7) We only consider the observe-then-download mode (Wang and Reinelt, 2010), namely an original image data is observed first, and only after the observation (acquisition of the image) is completed, the downlink tasks for transmitting it could be executed.

## 2.1. Components of D-SIDSP

An instance of D-SIDSP can be represented as

$$\mathcal{P} = \{St, Et, \mathcal{S}, \mathcal{G}, TW, OD, DT, Constraints\} \quad (1)$$

Let us now discuss in detail each of the components in the above representation of D-SIDSP.

(1) $[St, Et]$ is the scheduling horizon for D-SIDSP. We fixed the scheduling horizon at 24h from 2020/10/15 00:00:00 to 2020/10/16 00:00:00.

(2) $\mathcal{S} = \{s : |\mathcal{S}| = n_s\}$ is the set of all available earth observation satellites (EOSs) considered in the D-SIDSP. The minimum volume of each segmented image data is a main attribute of each $s$ considered in D-SIDSP. An $s$ in S can be represented as an 8-tuple

$$s = \{Id, \Omega, i, a, e, \omega, M_0, d_0\} \quad (2)$$

where $Id$ is the identifier of $s$ and $\{\Omega, i, a, e, \omega, M_0\}$ denotes the six orbital elements for $s$.

The value $d_0$ is the minimum volume of a segmented image data, which will be used to restrict segmentation of each of the original image data. We use the transmission duration for defining the minimum volume with its unit as second. As indicated earlier towards the end of the introduction section regarding notational convention, all attributes of $s$ will be represented in the format "$s.*$". For example, the identifier $Id$ of $s$ will be denoted by $s.Id$. Without further explanations, we follow this notational convention throughout this paper.

(3) $\mathcal{G} = \{g : |\mathcal{G}| = n_g\}$ is the set of all available ground stations (GSs) in D-SIDSP. Each $g$ is composed of a six-tuple:

$$g = \{Id, lat, lon, alt, \gamma, \pi\} \quad (3)$$

where $Id$ is the identifier of $g$, $\{lat, lon, alt\}$ denotes the location of $g$, and $\gamma$ and $\pi$ denote the maximum angle of roll and pitch of the antenna installed in $g$. These two angles restrict the visibility of GSs by the EOSs. Before receiving image data, a ground station will spend some time, called *the antenna set-up time*, to calibrate the antenna to EOS. We discuss *the details of antenna set-up time* later.

(4) $TW = \{tw : |TW| = n_{tw}\}$ indicates the set of all available visible time windows (VTWs) between GSs with EOSs during the scheduling horizon and is called the transmission windows. Here $n_{tw}$ means the number of all transmission windows. Each transmission window $tw$ is represented by a 5-tuple

$$tw = \{Id, g, s, s, e\} \quad (4)$$

where $\{Id, g, s\}$ confirm to the identity of $tw$ and $Id$ is the window identifier of $tw$. $g$ indicates the GS that $tw$ belongs to, $s$ denotes the visible EOS, and $s$ and $e$ respectively represent begin time and end time of the $tw$.

(5) $OD = \{od, |OD| = n_{od}\}$ indicates the set of all valid original image data available during the scheduling horizon and $n_{od}$ is the number of original image data. Each original image data $od$ in OD is represented by a 7-tuple

$$od = \{Id, sId, s, \omega, d, r, o\} \quad (5)$$

where $Id$ is the identifier of $od$ and $sId$ represents segmented image data. Note that $sId$ is valid only when $od$ represents a segmented image data, otherwise $sId$ is invalid. $s$ denotes the EOS that $od$ belongs to, $\omega$ indicates the priority of $od$, $d$ is the transmission duration of $od$, $r$ denotes the release time of $od$ and $o$ denotes the due time of $od$, which is related to its priority (Li et al., 2014). We use a piecewise function shown in (6) to calculate the due time.

$$gt.o = \begin{cases} 24 & gt.\omega \in [1,3] \\ 12 & gt.\omega \in [4,6] \\ 6 & gt.\omega \in [7,9] \\ 3 & gt.\omega = 10 \end{cases} \quad (6)$$

The unit of the due time is in hours. In addition, $[r, r + o]$ denotes the validity time window of $od$ where $r + o$ is the expiration date (time).

(6) $DT = \{dt : |DT| = n_{dt}\}$ is the set of all generated downlink tasks during the scheduling horizon. Each $dt$ in DT is composed of six elements. i.e.

$$dt = \{Id, b, e, d, tw, dSet\} \quad (7)$$

where $Id$ is the identifier of $dt$, $b$ and $e$ respectively denote the transmission begin moment and the transmission end moment of $dt$, $d$ indicates the processing time of $dt$, and $d = \sum_{od \in dSet} od.d$. Further, $tw$ represents the transmission window for executing $dt$, the structure of which is defined by the function (4), and $dSet$ indicates the set of (segmented) image data transmitted in $dt$. The elements of $dSet$ is defined as the function (5). In addition, all image data in $dSet$ must belong to the same EOS. i.e. $tw.s = od.s \ \forall od \in dSet$.

Let us now examine the constraints. Most existing research works on SIDSP (Malladi et al., 2016, Liu et al., 2017, Karapetyan et al., 2015) considered time restrictions as an important constraint. For the D-SIDSP, there are four types of time constraints, which are *the visible time constraint*, *the work time constraint, the logical time constraint,* and *the antenna set-up time constraint*.

**The visible time constraint** states that every downlink task should be in the corresponding visible time window (VTW). For the downlink task $dt$, this constraint can be described as

$$\begin{cases} dt.b \geq dt.tw.s \\ dt.e \leq dt.tw.e \end{cases} \quad (8)$$

The processing time of each downlink task should be bigger than the minimum processing time and less than the length of the corresponding transmission window. Thus, for the downlink task $dt$, **the work time constraint** can be expressed as

$$\begin{cases} dt.d \geq dt.tw.s.d_0 \\ dt.d \leq dt.tw.e - dt.tw.s \end{cases} \quad (9)$$

Note that we are allowing multi-EOS in D-SIDSP and hence the minimum processing time should correspond to the right EOS.

**The logical time constraint** is a hard constraint that cannot be violated and reflects that the image data should be generated at first and only after that the data can be transmitted, as mentioned in the assumption (7). In other words, for an (original/segmented) image data, its release time should be earlier than its transmission begin time. On the other hand, it also should be transmitted before it is overdue. For an image data $od$, let $dt$ be the downlink task to receive $od$, then the logical constraint can be expressed as

$$\begin{cases} dt.s == od.s \\ od.r \leq dt.b \\ od.o + od.r > dt.b \end{cases} \quad (10)$$

**The antenna set-up time** is used to calibrate the antenna of the corresponding EOS. As mentioned in the assumption (2), one GS can receive image data only from one EOS in any given time. There is an interval time, the antenna set-up time ($\sigma_{s_i \to s_j}^g$), used to turn the antenna from one EOS to another, and during $\sigma_{s_i \to s_j}^g$ no transmissions can occur (Marinelli et al., 2011). $\sigma_{s_i \to s_j}^g$ depends on rotation angle and rotation speed and is given by

$$\sigma_{s_i \to s_j}^g = \frac{|Angle_{s_i} - Angle_{s_j}|}{rev_g} \quad s_i, s_j \in S, g \in G \quad (11)$$

where $Angle_{s_i}$ and $Angle_{s_j}$ are the angles when $g$ receives the image data from $s_i$ and $s_j$ respectively.

Many earlier works (Du et al., 2019, Zhang et al., 2019, Zufferey et al., 2008) viewed $\sigma_{s_i \to s_j}^g$ as a constant value. Fabrizio (Marinelli et al., 2011) discussed different values of $\sigma_{s_i \to s_j}^g$ under different processing times of ground stations. They observed that $\sigma_{s_i \to s_j}^g$ can be ignored when the processing time for ground stations are bigger than 10 minutes. Therefore, we assume $\sigma_{s_i \to s_j}^g$ as a constant $\sigma = 60\ s$. For two adjacent downlink tasks, $dt_i$ and $dt_{i+1}$, with $dt_i$ is executed before $dt_{i+1}$, *the antenna set-up time constraint* can be described as

$$dt_{i+1}.b - dt_i.e \geq \sigma \quad (12)$$

Note that these two downlink tasks have to belong to the same GS but not the same EOS. That is $dt_i.tw.g = dt_{i+1}.tw.g$ and $dt_i.s \neq dt_{i+1}.s$.

In addition, there are two types of performance constraints, *completed transmission constraint* and *segmented constraint*, specially handled in D-SIDSP. To facilitate the definition of these two constraints, some additional notations are introduced below.

- $od$     An original image data
- $SD$     The set of all segmented image data from $od$ and all of them are transmitted
- $sd$     An arbitrary segmented image data in the set $SD$, and it is defined as the function (5)

**The completed transmission constraint** is used to restrict that all segmented image data from the same original image data should be transmitted completely, which is consistent with the assumption (3). This constraint can be described as

$$\sum_{sd \in SD} sd.d = od.d \qquad (13)$$

Note that the basic assumption regarding the function (13) is that the original image data ($od$) is scheduled to be transmitted. So, if $od$ is not scheduled to be transmitted, the function (13) is invalid.

**The segmented constraint** is used to restrain the segmentation process for every original image data. It is apparent that too small segmentation is bound to bring difficulties to image processing on the ground and also not conducive to the operation of the satellite system. So, we do not consider over segmentation of the original image data and set the minimum size of image data as defined in the function (2). Therefore, this constraint can be described as

$$sd.d \geq od.s.d_0 \quad \forall sd \in SD \qquad (14)$$

## 2.2. A bi-objective optimization model

Based on detailed problem description above, three classes of decision variables, $x_i$, $y_i^j$ and $dt_j.b$, are considered to build D-SIDSP. In addition, to facilitate the development of the model, some additional notations and symbols are introduced below.

- $od_i$     An arbitrary original image data
- $tw_j$     An arbitrary transmission window
- $dt_j$     An arbitrary downlink task
- $x_i$      A binary variable, $x_i = 1$ represents $od_i$ is scheduled to be transmitted, otherwise $od_i$ is not considered.
- $y_i^j$    A non-negative continuous variable which denotes the part of $od_i$ transmitted in $tw_j$ and belongs to [0,1]. $y_i^j > 0$ represents $od_i$ is scheduled to be transmitted in $tw_j$, otherwise it is not scheduled.
- $dt_j.b$   A non-negative continuous variable, as mentioned above, $dt_j.b$ represents the transmission begin moment of a downlink task $dt_j$.
- $OD_s$     The set of all image data in EOS $s$
- $TW_s$     The set of all available transmission windows for EOS $s$

Transmitting as much image data as possible is the original intention of (D-)SIDSP. Several optimization objective functions such as, maximize transmission revenue (Karapetyan et al., 2015, Marinelli et al., 2011), maximize downlink duration (Zhang et al., 2019) and minimize transmission failure rate (Du et al., 2019, Chang et al., 2021a), were adopted in many existing research works. Without loss of generality, we set minimizing transmission failure rate as one of the optimization objective in our study.

The first objective, $f_1(\mathcal{P})$, is ***the image data transmission failure rate (FR)***. FR is calculated by considering the priority of all original image data and is defined as

$$f_1(\mathcal{P}) = 1 - \frac{\sum_{od_i \in OD}(x_i \times od_i.\omega)}{\sum_{od_i \in OD}(od_i.\omega)} \quad (15)$$

where $\sum_{od_i \in OD}(x_i \times od_i.\omega)$ denotes the total profit of transmitted image data, and $\sum_{od_i \in OD}(od_i.\omega)$ is calculated under the assumption that all of the original image data corresponding to $i$ is transmitted. By the way, $f_1(\mathcal{P})$ is normalized so that it belongs to the interval [0,1].

As mentioned earlier, the observation (imaging) capability is significantly superior to transmission/receiving capability. In other words, the available transmission windows are not enough for receiving all of the image data that the EOSs under consideration is able to acquire. Because of this scarcity of transmission windows, the optimization objective, called the load-balance degree of remote-tracking antennas proposed in (Du et al., 2019) is not very efficient. But according to their optimization objective, we would like to define a new optimization objective, ***the service-balance degree (SD)*** of EOSs, to evaluate the service balance among all EOSs, and the SD is the deviation of utilization ratio among available transmission windows of EOSs. In addition, $f_2(\mathcal{P})$(SD) considers work duration of downlink tasks for every EOS and can be defined as

$$f_2(\mathcal{P}) = \frac{\sum_{s \in S}(1 - UR_s)}{n_s} \quad (16)$$

where $UR_s$ represents the utilization ratio of transmission windows of EOS $s$ and defined as the function (17). $(1 - UR_s)$ denotes the distance for the utilized ratio of transmission windows for EOS $s$ from the absolute utilized ratio, which is equal to 1. In addition, $f_2(\mathcal{P})$ is a dimensionless variable and it belongs to the interval [0,1]. Here,

$$UR_s = \left.\sum_{tw \in TW_s} \left(\frac{dt^{tw}.d}{tw.e - tw.s}\right)\right/ |TW_s| \qquad (17)$$

where $dt^{tw}$ indicates the downlink task according to the transmission window $tw$. Note that, if $tw$ is not used to generate a downlink task, $dt^{tw}.d = 0$. $|TW_s|$ denotes the number of available transmission windows for EOS $s$.

Thus, we have two optimization objectives, *the image data transmission failure rate and the service-balance degree*, for D-SIDSP. These two objectives are two metrics that measure the performance of scheduling from different perspectives and irreconcilable in satellite image data downlink scheduling, so the simultaneous optimization of them is reasonable and possible. Thus, our objective is to minimize the two criteria defined above. i.e.

$$\min F(\mathcal{P}) = \{f_1(\mathcal{P}), f_2(\mathcal{P})\} \qquad (18)$$

Thus, the D-SIDSP is a bi-objective discrete optimization problem, which has disconnected nondominated solution sets (Kidd et al., 2020). Now, we will analyze the constraints of the D-SIDSP. Our primary constraints are summarized below, followed by a brief discussion of each of them.

$$x_i \leq 1 \qquad (19)$$

$$x_i == 1 \Rightarrow \begin{cases} y_i^j \times od_i \geq od_i.s.d_0 \\ \sum_{j=1}^{n_{tw}} y_i^j == 1 \end{cases} \qquad (20)$$

$$y_i^j > 0 \Rightarrow \begin{cases} dt_j.tw.s == od_i.s \\ od_i.r \leq dt_j.b \\ od_i.r + od_i.o > dt.b \end{cases} \qquad (21)$$

$$\left.\begin{array}{r} \sum_{i=1}^{n_{od}} y_i^j > 0 \text{ and } \sum_{i=1}^{n_{od}} y_i^k > 0 \\ dt_j.tw.g == dt_k.tw.g \\ dt_j.tw.s \neq dt_k.tw.s \\ dt_j.b < dt_k.b \\ \nexists \sum_{i=1}^{n_{od}} y_i^l > 0 \\ dt_l.tw.g == dt_k.tw.g \\ dt_l.b \in [dt_j.e, dt_k.b] \end{array}\right\} \Rightarrow dt_k.b - dt_j.e \geq \sigma \qquad (22)$$

$$\sum_{i=1}^{n_{od}} y_i^j > 0 \Rightarrow \begin{cases} dt_j.d \geq dt_j.g.d_0 \\ [dt_j.b, dt_j.e] \subseteq [dt_j.tw.s, dt_j.tw.e] \end{cases} \qquad (23)$$

$$y_i^j \in [0,1] \qquad (24)$$

$$x_i \in \{0,1\} \qquad (25)$$

The constraint (19) indicates that each of the original image data can be acquired at most once.

The constraint (20) denotes the segmentation constraint, defined as the function (14), and the completed transmission constraint, defined as the function (13). The constraint (21) denotes the logical time constraint, defined as the function (10). The constraint (22) expresses the antenna set-up time constraint as defined in the function (12). In addition, the first part of the constraint (22) indicates $dt_j$ and $dt_k$ are two arbitrary adjacent downlink tasks belong to a same ground station but not the same EOS. The constraint (23) denotes the work time constraint, defined as the function (9), and the visible time constraint, defined as the function (8). Note that, since the minimum work time of a downlink task equals the minimum size of the image data, it will certainly satisfy the first equation in constraint (23). In addition, the constraints (24) and (25) reflect the available value of $y_i^j$ and $x_i$. It may be noted that the way we represent the constraints is not suitable for directly using mixed integer programming solvers. Our goal is to develop a memetic algorithm where we only require capabilities for evaluating the objective function value of a trial solution and verifying feasibility of the trial solutions. For these purposes, our model representation is adequate.

## 3. An adaptive bi-objective memetic algorithm

Memetic computing/algorithm (MC/MA) usually (Neri and Cotta, 2012) consists of an evolutionary framework and a set of local search algorithms, which are activated within the generation cycle. This algorithmic framework has been successfully used in solving various combinatorial optimization problems.

We design an adaptive bi-objective memetic algorithm, called ALNS+NSGA-II, to solve D-SIDSP. ALNS+NSGA-II combines an adaptive large neighborhood search algorithm (ALNS) and a nondominated sorting genetic algorithm II (NSGA-II), and the general framework is outlined in the flow chart of Figure 2.

ALNS is used as the local search algorithm within ALNS+NSGA-II to breed offspring solutions. ALNS can search for good quality solutions faster and has been adopted in some of the earlier studies (Liu et al., 2017, He et al., 2018, He et al., 2019, Kadziński et al., 2017, Wu and Che, 2019, Wu and Che, 2020). The basic structure of ALNS is constructed around two loops. The inner loop is a local search process, which consists of some "Destroy" operators and some "Repair" operators, while the outer loop uses some control mechanisms to guide the search process. A score and a weight are assigned to each operator, and an adaptive layer is included to update the weights and scores of different operators according to their performance in searching for better solutions.

NSGA-II is used as the evolutionary mechanism for ALNS+NSGA-II to achieve the (approximate) Pareto frontier faster. Deb et al. (Deb et al., 2002) improved NSGA (Srinivas and Deb, 1995) and proposed NSAG-II in 2002, which is one of the best known evolutionary multi-objective optimization algorithms (Gong et al., 2009). NSGA-II can obtain the Pareto frontier faster for a fast nondominated sorting approach, and they also proposed a fast-crowded distance estimation and a simple crowded comparison operator to overcome the difficulty that evolutionary algorithms are depended on parameters.

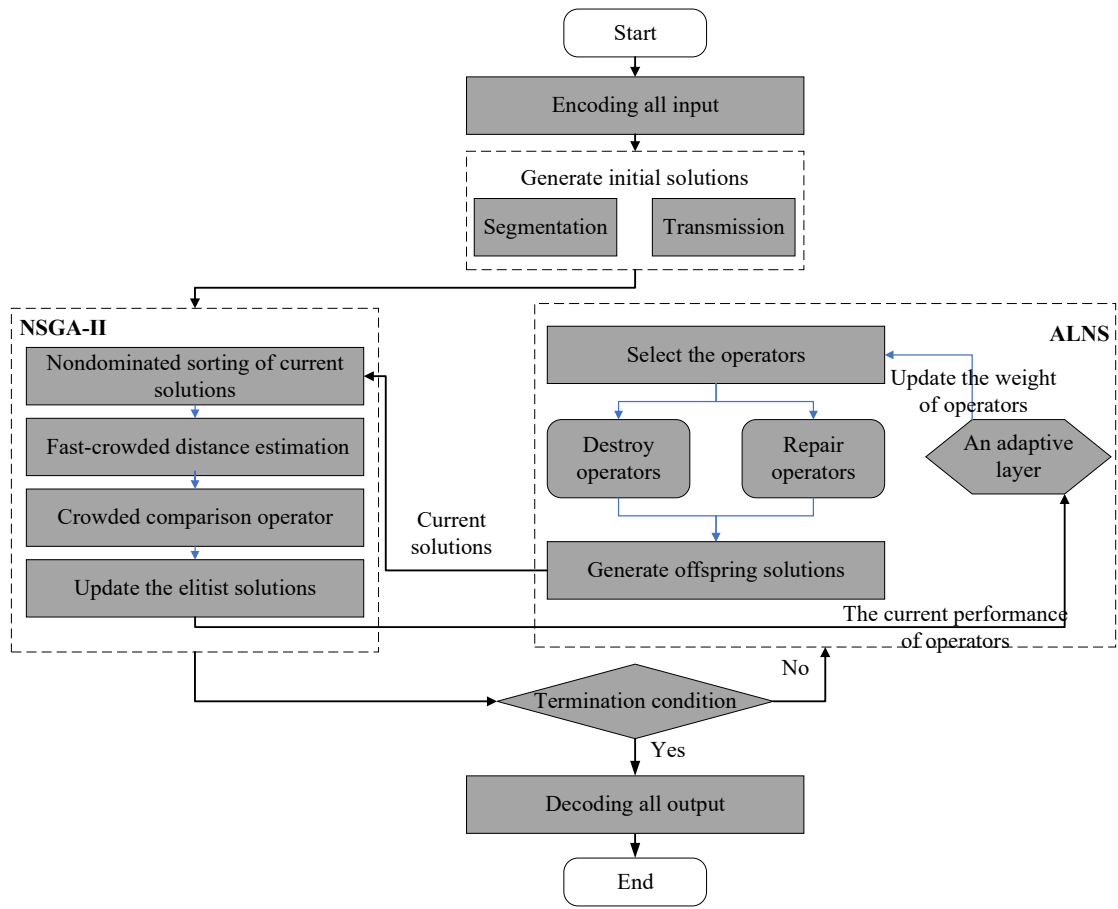

Figure 2 The basic flow of ALNS+NSGA-II

In addition, our algorithms will adopt the well-known box-method (Hamacher et al., 2007), which is based on $\varepsilon$-constraint method, to guide the evolution in the non-dominated space.

## 3.1. A hybrid encoding

Recall that there are three types of decision variables, $x_i$, $y_i^j$ and $dt_j.b$, in our model. The variables $x_i$ and $y_i^j$ are used directly, whereas as $dt_j.b$ is defined (and maintained) in terms of another variable $z_j$ and it belongs to the interval [0,1]. The relationship between $z_j$ and $dt_j.b$ is given by

$$dt_j.b = dt_j.tw.s + z_j \times (dt_j.tw.e - dt_j.tw.s) \tag{26}$$

Also note that $x_i$ is a binary variable, while $y_i^j$ and $z_j$ are continuous non-negative variables. So, a hybrid encoding scheme is proposed to express the solutions of D-SIDSP. An example of this is shown as Figure 3.

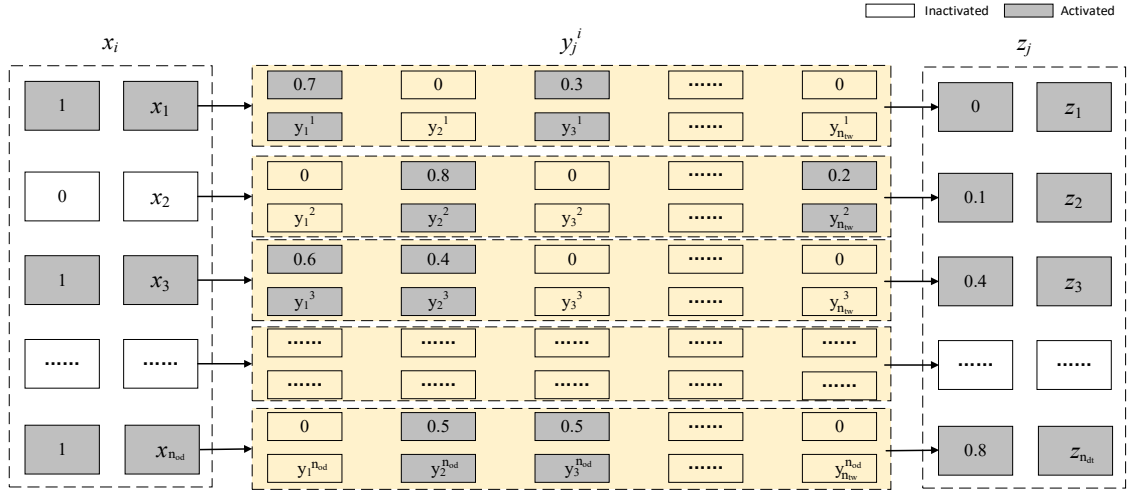

Figure 3 An example of hybrid encoding

As illustrated in Figure 3, the dark rectangle indicates the corresponding code is valid, while the light one indicates the corresponding code is invalid. The coding for $x_i$ is obvious, the value equals 1 means the corresponding original image data is scheduled to be transmitted, otherwise it is not. As mentioned above, $y_i^j$ denotes the part of $od_i$ transmitted during $tw_j$ and it belongs to [0,1]. If $y_i^j > 0$, it represents that $od_i$ is scheduled to be transmitted in $tw_j$, otherwise it is not. Note that the available transmission windows for every original image data are distributed in the time span determined by the expiration date (time) of the original image data. The value of $z_j$ reflects the transmission begin moment in the corresponding transmission window. As mentioned earlier, the available transmission windows are not enough for receiving all possible image data, so all available transmission windows are likely to be used. In addition, the value of each decision variable is also marked in the figure.

## 3.2. A random heuristic greedy algorithm (RHGA)

Since ALNS is not sensitive (Ropke, 2007) to initial solution, we can construct it randomly. In addition, solutions obtained by heuristic greedy algorithms are always feasible. Therefore, a random heuristic greedy algorithm (RHGA) is proposed as the initialization algorithm. The pseudocode for RHGA is shown as Algorithm 1. In addition, to facilitate a better understanding of the pseudocode l, some additional notations and symbols are introduced and explained below.

- $od$    An arbitrary original image data
- $SD$    The segmentation scheme of $od$
- $DT$    The transmission scheme
- $OD$    A set of all original image data
- $TW$    A set of all transmission windows
- $\mathbb{s}$    An initial solution. Note that, it is expressed in the hybrid coding
- $ATW$    The set of available transmission windows for transmitting $od$
- $atw$    A valid transmission window
- $sd$    A segmented image data

---
Algorithm 1 : A random heuristic greedy algorithm (RHGA)

**Input:** $OD$ and $TW$
**Output:** $\mathbb{s}$

---
1:   Encode $OD$ and $TW$
2:   **Traverse**------ Generate the transmission scheme
3:    $od \leftarrow OD$ --- **Greedy**
4:     **Repeat**---- Confirm $\phi$ and update $\psi$
5:      $SD \leftarrow$ **ODCS** ();
6:      $DT \leftarrow$ **SRS** ();
7:     **Until** $od$ is scheduled or cannot be transmitted
8:    Update $TW$ according to $DT$.
9:   **Until** $OD$ is visited completely.
10:   Calculate $\{f_1(\mathcal{P}), f_2(\mathcal{P})\}$.
11:   **Return** $\mathbb{s}$.

---

Note that the hybrid encoding method discussed earlier is adopted in line 1:. Lines 2: to 9: are used to confirm segmentation scheme for all original image data and generate downlink tasks. Among them, all original image data will be selected one by one in a greedy way (the line 3:), and before the selection, all original image data are ranked by the given information1. Then, based on the newest available transmission windows ($TW$) and all fixed and scheduled original image data, cut and stock the selected original image data to confirm the segmentation scheme by ***ODCS*** (), defined as Algorithm 2. After that, generate/update the transmission scheme by ***SRS*** (), defined as Algorithm 3.

| Algorithm 2 : ***ODCS*** () |
| --- |
| **Input:** The newest set $TW$ and $od$ |
| **Output:** $SD$ |
| 1:    $ATW \leftarrow TW$ according to the constraint (10) and (14). |
| 2:    Calculate the sum of work time of $ATW$, if the sum is less than the transmission duration of $od$, **Return** false directly. |
| 3:    **Repeat**-----Confirm $SD$ |
| 4:        $atw \leftarrow ATW$ to transmit $od$----**Greedy** |
| 5:        Confirm $SD$ according to the constraint (13) and (14). |
| 6:        If all constraints are satisfied and $od$ has been transmitted, and end this repeat. |
| 7:    **Until** $ATW$ is visited completely. |
| 8:    **Return** $SD$ |

As mentioned above, ***ODCS*** () is designed to cut and stock the selected original image data. All valid transmission windows are updated based on the newest available transmission windows. If there is a valid segmentation scheme ($SD$) for the selected original image data, break ***ODCS*** () directly, otherwise it will traverse through all valid transmission windows. In addition, ***ODCS*** () does not confirm the final transmission scheme for the selected original image data, but only obtain a feasible segmentation scheme, namely a set of segmented image data.

| Algorithm 3 : ***SRS*** () |
| --- |
| **Input:** $SD$, $TW$ and $DT$ |
| **Output:** The updated $DT$ |
| 1:    **Fix** all scheduled image data --- **Greedy**. |
| 2:    **Repeat** ------ Confirm $DT$ |
| 3:        $sd \leftarrow SD$ |
| 4:        $atw \leftarrow TW$ according to the constraint (10) for transmitting $sd$ |
| 5:        **Generate/update** $DT$ based on used $TW$ according to the constraint (8), (9) and (11). |
| 6:    **Until** $SD$ is transmitted completely |
| 7:    **Return** $DT$. |

***SRS*** () is proposed to confirm transmission window for all segmented image data and then generate/update downlink tasks. All segmented image data from the same original image data have to be transmitted completely, otherwise it is marked 'fail'. In addition, the dichotomization method is adopted to confirm the transmission completion moment of every downlink task in the line 5:, which will improve the speed of searching, to some extent.

---

1 To generate the initial solutions, the given information is random.

### 3.3. "Destroy" operators and "Repair" operators

Let us now consider two operators - "Destroy" and "Repair" - to design the adaptive large searching neighborhood (ALNS), for D-SIDSP.

"Destroy" operators are used to change the elements (original image data) of ALNS by removing some transmitted original image data, while "Repair" operators are used to rank all available elements of ALNS by different available information to change ALNS. Also a taboo bank with adaptive size is proposed to improve the efficiency of these operators. Considering the performance of each operator, an adaptive layer is designed in section 3.4 to select the operators.

**3.3.1. "Destroy" operators**

Since there are two primary objects for processing, *image data*, and *downlink tasks*, in D-SIDSP, we design two kinds of "Destroy" operators for image data and downlink tasks respectively. The first kind of "Destroy" operators, **RD-Destroy**, **PD-Destroy**, **DD-Destroy** and **CD-Destroy**, remove some original image data directly, while the second kind of "Destroy" operators, **RT-Destroy**, **PT-Destroy**, **WT-Destroy** and **CT-Destroy**, first chose some downlink tasks and then remove all (segmented) image data transmitted in them. If a part of an original image data is removed, then all parts of that original image data will be removed. All removed original image data are saved into a set called taboo bank $B$ with a given size $|B|^2$, which represents all of the image data that are not to be considered when "Repair" operators execute. $B$ is empty before removing image data and filling $B$ to its full capacity is the termination criterion for "Destroy" operators. Also, all unscheduled image data is saved in a set denoted by $F$. The image data in $F$ but not in $B$ will be considered by "Repair" operators to produce offspring solutions. Let us now discuss each type of "Destroy" operators in detail.

---

[2] The size of the taboo bank $B$ is adaptive and belongs to the interval [0,0.2]. In addition, we will analyze the effect of the different sizes of $B$ on the efficiency of ALNS+NSGA-II in the simulation experiment.

**RD-Destroy:** This operator removes some scheduled original image data from the given solution randomly.

**PD-Destroy:** This operator ranks all scheduled original image data by their priority ascending and then deletes them one by one, which represents the scheduled original image data with lower priority will be removed from the given solution in priority.

**DD-Destroy:** This operator ranks all scheduled original image data by their transmission duration descending, which denotes that the scheduled original image data spent more time to transmit will be removed from the given solution in priority.

**CD-Destroy:** The guiding information of this operator considers the workpiece congestion defined in (Chang et al., 2020). For example, for the original image data $od_i$ the workpiece congestion ($GI\_CD$) is defined as

$$GI\_CD(od_i) = \sum_{j=1}^{n_{od}} NoD(od_j.\omega \times d_{ij}) \tag{27}$$

where $d_{ij}$ denotes the conflict distance between $od_i$ and $od_j$. If there is an irreconcilable conflict between them, i.e. either-or conflict, $d_{ij} = 1$; if there is a reconcilable conflict, $d_{ij} = 0.5$; else if there is no conflict, $d_{ij} = 0$. $NoD()$ represents the dimensionless processing value defined as

$$NoD(x_i) = \frac{1}{\exp(1 - x_i / \max_{j=1,\cdots,n} x_j)} \tag{28}$$

where $x_i$ is an independent variable, $\max_{j=1,\cdots,n} x_j$ represents the maximum value in the set $\{x_j | j = 1, \cdots, n\}$ and $n$ indicates the size of the set.

Note that, this operator will rank scheduled original image data by their workpiece congestion ascending (Chang et al., 2020).

**RT-Destroy:** This operator selects some downlink tasks randomly, and then remove some (segmented) image data transmitted in them randomly. This way, all other segmented image data will be removed, from the same original image data with removed segmented image data.

**PT-Destroy:** The guiding information of this operator is similar to that of **PD-Destroy** and considers the priority of all scheduled image data. For the downlink task $dt$, the guiding information ($GI\_PT$) is defined as

$$GI\_PT(dt) = \sum_{od \in dt.dSet}(od.\omega) \tag{29}$$

where $dt.dSet$ denotes the set of image data transmitted in $dt$. In addition, this operator ranks all downlink tasks by $GI\_PT$ descending and then remove the (segmented) image data in them one by one. Note that all (segmented) image data in every downlink task is also sorted by their priority descending.

**WT-Destroy:** This operator ranks all downlink tasks by their processing time descending, and considers the downlink tasks with longer processing time in priority. All of the (segmented) image data in every downlink task is sorted by their transmission duration descending and removed one by one.

**CT-Destroy:** The guiding information of this operator is similar to that of **CD-Destroy**. and considers the workpiece congestion of all scheduled image data. For the downlink task $dt$, the guiding information ($GI\_CT$) is

$$GI\_CT(dt) = \sum_{od \in dt.dSet}(GI\_CD(od)) \qquad (30)$$

where $GI\_CD(od)$ represents the workpiece congestion of image data $od$ defined by (27).

Note that, this operator will rank all downlink tasks by their workpiece congestion ascending (Chang et al., 2020).

### 3.3.2. "Repair" operators

All unscheduled original image data is saved in a set denoted by $F$. The original image data in $F$ which is not in $B$ could be selected and inserted into a given solution to produce an offspring solution. RHGA is adopted to schedule all selected original image data into the given solution. Note that, both the segmentation scheme and the transmission scheme are rescheduled based on scheduled original image data. We consider four different "Repair" operators and the difference between them is their different guiding information for sorting the unscheduled original image data.

**R-Repair**: This operator selects some unscheduled original image data in $F$ but not in $B$ randomly and inserts them into the given solution.

**P-Repair**: This operator uses an ascending arrangement of selected original image data by the guiding information which denotes the unscheduled original image data with higher priority will be considered in priority.

**S-Repair**: This operator ranks all unscheduled original image data by the number of their available transmission windows (ATWs) in an ascending way, which presents the unscheduled original image data with fewer ATWs will be considered in priority.

**C-Repair**: The guiding information of this operator is similar to that of **CD-Destroy**, while **C-Repair** will consider unscheduled original image data with less workpiece congestion in priority.

### 3.4. An adaptive layer and termination criterion

Each operator has a score and a weight. The score depends on the performance of observation scheduling, and the weight is updated according to the score. The four scores considered in our algorithms are defined as follows.

- $\sigma_1$    If the new solution dominates all current solutions
- $\sigma_2$    If the new solution dominates one of the current non-dominated solutions
- $\sigma_3$    If the new solution is located on the current Pareto frontier
- $\sigma_4$    If the new solution is dominated by one of the current non-dominated solutions

At the end of every iteration, the weights of each of the operators are updated according to the formula (31).

$$\omega_i^\alpha = (1-\lambda)\omega_i^\alpha + \lambda \frac{\pi_i^\alpha}{\sum_{j=1}^{|I_\alpha|}\pi_j^\alpha} \quad 1 \leq i \leq |I_\alpha| \tag{31}$$

where, $\alpha$ belongs to {Destroy, Repair} denotes the type of the operator, and $|I_\alpha|$ represents the number of operators of the corresponding type. The values $\pi_i^\alpha$ and $\omega_i^\alpha$ respectively denote the score and weight of the $i^{th}$ operator and $\lambda \in [0,1]$ is a reaction factor that controls how sensitive the weights are to changes in the performance of the operators. A value of 0 means that the weights remain unchanged, while a value of 1 implies that the historic performance has no impact. Thus, the weight only depends on the current score.

The roulette wheel mechanism is used to choose appropriate operators[3]. The utilization rate $(r_i^\alpha)$ is calculated using the equation:

$$r_i^\alpha = \frac{\omega_i^\alpha}{\sum_{j=1}^{|I_\alpha|}\omega_j^\alpha} \quad 1 \leq i \leq |I_\alpha| \tag{32}$$

The maximum number of iterations, denoted as $MaxIter$, is the only termination criterion for ALNS+NSGA-II. $MaxIter$ is given before every evolution cycle.

---

[3] In the first iteration, all operators will be used in the equal possibility, then they will be chosen by the roulette wheel mechanism.

# 4. Simulation experiments

In this section, we carry out experimental analysis of our algorithm ALNS+NSGA-II for solving D-SIDSP, from various points of views. In addition, we also generate various test instances to be used in our experiments. All algorithms are coded in C#, using Visual Studio 2013, and tested on a laptop with intel(R) Core (TM) i7-8750H CPU @ 2.2GHz and 16 GB RAM. Some general parameter settings for ALNS+NSGA-II are shown in Table 1.

Table 1 General parameters

| Parameter | Meaning | Value |
|---|---|---|
| $NS$ | The population size of all preserving solution | 100 |
| $NA$ | The population size of archive solutions | 100 |
| $MaxIter$ | The maximum number of iterations | 200 |
| $TR$ | The rate of the size of taboo bank to the scheduled ground targets | [0,0.2] |
| $\lambda$ | The value of the reaction factor to control update the weight of operators | 0.5 |
| $\sigma_1$ | If the new solution dominates all current solutions; | 30 |
| $\sigma_2$ | If the new solution dominates one of the current non-dominated solutions; | 20 |
| $\sigma_3$ | If the new solution on the current Pareto frontier; | 10 |
| $\sigma_4$ | If the new solution is dominated by one of the current non-dominated solutions. | $\{0,1\}$[4] |

## 4.1. Simulation instances

There are no publicly available benchmark instance for the SIDSP. Many researchers designed their own test instances based on considerations derived from the real world situations (Karapetyan et al., 2015, Li et al., 2014, Xiao et al., 2019). We also followed this strategy in generating test instances, including the original image data, ground stations, and earth observation satellites.

**4.1.1. Ground stations**

The China remote sensing satellite ground station (RSGS) became operational in 1986 with the completion of a ground station in Miyun (40°N/117°E), near Beijing. In recent years, Kashi Station (39°N/76°E), located in Xinjiang Province in western China, and Sanya Station (18°N/109°E), located in southern China's Hainan Province, have been constructed to expand satellite operations and reception coverage area (Guo et al., 2012). By the end of 2016, the China

---

[4] To avoid being in the local optimum all the time, the value of $\sigma_4$ equals 1 with a small possibility, 0.1. That is, a worse solution will be accepted with the small possibility.

Remote Sensing Satellite North Polar Ground Station (CNPGS, 67°N/ 21°E), China's first overseas land satellite receiving station, was put into trial operation near Kiruna, Sweden (Na, 2016). Therefore, we used these four ground stations, 3 normal ground stations and 1 polar station, in our research work and experimental analysis of algorithms.

**4.1.2. Earth observation satellites**

In addition, we select ten state-of-the-art Chinese low orbit EOSs (LEOSs) from the satellite database in the AGI Systems Tool Kit (STK) 11.2, the three EOSs are from the series of "Gao Fen" satellites, the four EOSs are from the series of "Super View" satellites and others are from the series of "Earth Resources" satellites. The basic attributes of them are shown in Table 2. As mentioned earlier, the scheduling horizon is 24 hours, from 2020/10/15 00:00:00 to 2020/10/16 00:00:00. Note that, all completed data about orbits of EOSs, location of ground stations and TWs can be found in our data files (available on request from the first author). In addition, the time of all data is cumulative seconds based on 2020/10/15 00:00:00.

Table 2 The basic attributes of EOSs

| Name | $Id$ | $\Omega(km)$ | $i$ | $a$ | $e$ | $\omega$ | $M_0$ | $d_0(s)$ |
|---|---|---|---|---|---|---|---|---|
| GF0101 | 1 | 7145.08 | 0.001 | 98.55 | 359.06 | 152.17 | 265.39 | 30 |
| GF0201 | 2 | 7011.57 | 0.002 | 97.83 | 2.89 | 98.15 | 257.45 | 30 |
| GF0601 | 3 | 7020.45 | 0.002 | 97.99 | 6.87 | 56.94 | 94.33 | 30 |
| SV01 | 4 | 6901.65 | 0.002 | 97.43 | 1.01 | 124.24 | 242.68 | 10 |
| SV02 | 5 | 6894.39 | 0.001 | 97.54 | 11.87 | 128.22 | 90.39 | 10 |
| SV03 | 6 | 6883.14 | 0.000 | 97.51 | 5.98 | 341.26 | 106.70 | 10 |
| SV04 | 7 | 6884.95 | 0.004 | 97.51 | 6.14 | 92.52 | 195.65 | 10 |
| ZY02C | 8 | 7143.90 | 0.002 | 98.64 | 341.91 | 57.55 | 186.17 | 60 |
| ZY3 | 9 | 6875.80 | 0.001 | 97.41 | 0.79 | 59.20 | 71.87 | 60 |
| ZY0104 | 10 | 7145.08 | 0.001 | 98.55 | 359.06 | 152.17 | 265.39 | 60 |

**4.1.3. Original image data**

Considering 3 normal ground stations and 1 polar station, there are about 1500 seconds of downlink time (not considering overlaps) for each EOS each day (about 14.5 orbits), so there is about 100 seconds in each orbit for each EOS to observe, which serves as our guiding information for generating the original image data. We design three types of realistic instances based on the types of the adopted ground stations, Normal distribution (ND), Polar distribution (PD) and Mixed distribution (MD). ND only considers 3 normal ground stations, PD considers the unique polar ground station, while MD considers all ground stations. The number of original image data of every distribution is shown in Table 3.

Table 3 Three types realistic instances

| Distribution name | Ground stations | Number of original image data |
|---|---|---|
| ND | Normal stations | [50, 500], with an increment step of 50 |
| PD | Polar station | [50, 500], with an increment step of 50 |
| MD | Normal stations + Polar station | [100,1000], with an increment step of 100 |

The priority of all original image data is uniformly generated from [1, 10], and the transmission duration of all original image data belongs to [10,200], with unit as second. In particular, for the three "Gao Fen" satellites, the transmission duration of the original image data belongs to [60,120]. For the four "Super View" satellites, this value belongs to [10,60], and for other "Earth Resources" satellites, it belongs to [120,200]. In addition, the release time of each original image data is uniformly distributed in the time interval from 2020/10/14 00:00:00 to 2020/10/16 00:00:00, and the due time of each original image data is calculated by the formula (6). According to the release time and the due time, we can filter all valid original image data in the scheduling horizon (2020/10/15 00:00:00 to 2020/10/16 00:00:00). All original image date will be available in our data files, which can be obtained from the first author.

## 4.2. The efficiency of "Segment & Rearrange"

Allowing segmentation of the original image data and permitting its rearranged transmission is the primary research motivation in this paper and distinguishes our work from research papers on traditional SIDSP. Thus we will analyze the effect of each components in the definition of D-SIDSP. We first would like to consider the experimental group evaluating the efficacy of "Segment & Rearrange", in which all original image data can be segmented and then transmitted without considering the acquisition order. In addition, in order to better explain the efficiency, three control groups are considered.

(1) In the first control group, denoted by "Segment & FOFD", all original image data can be segmented, but must be transmitted as "First observation, First downlink (FOFD)" as in (Wang et al., 2011, Wang and Reinelt, 2010). Note that, under FOFD, original image data could be deleted without downlink directly.

(2) The second control group, denoted by "Unsegment & Rearrange", considers that the original image data cannot be segmented, but can be transmitted without considering the acquisition order, which is similar to the focus in (Karapetyan et al., 2015).

(3) In the last control group, denoted by "Unsegment & FOFD", the original image data cannot be segmented and must be transmitted one by one as per their release time ascending.

The final objective values of the elitist solutions and the value of Hypervolume (HV) are considered as performance indicators to compare the efficiency of ALNS+NSGA-II under different groups. HV is calculated by the Hypervolume by slicing objectives (HSO) (Bradstreet et al., 2008). The compared results are shown in Table 4. Where $HV$ denotes the value of HV[5], $\overline{v_1}$ and $\overline{v_2}$ represent the average final value of FR and SD on the Pareto frontier respectively.

(1) The average value of FR and SD obtained by ALNS+NSGA-II under "Rearrange", "Segment & Rearrange" and "Unsegment & Rearrange", are significantly less than those under "FOFD", other two control groups, for all simulation instances. Note that the smaller the value of FR, the larger is the transmission revenue, and the smaller the value of SD, the larger the fully utilized transmission windows and the more evenly the EOSs are served. That is, "Rearrange" not only make EOSs transmit more image data to ground stations, it also ensures that the transmission windows are used as much as possible, and every EOS gets more balanced service.

(2) The average value of FR and SD obtained by ALNS+NSGA-II under "Segment", "Segment & Rearrange" and "Segment & FOFD", are not consistently better than those under "Unsegment", other two control groups, toward all simulation instances. Especially, "Segment & FOFD" is even worse than "Unsegment & FOFD" occasionally. But "Segment & Rearrange" showed consistently better performance than "Unsegment & Rearrange".

(3) The values of HV under "Rearranger" is also significantly and consistently larger than that under "FOFD", while the values of HV under "Segment" is actually similar to that under "Unsegment". But combining "Segment" with "Rearrange", that is "Segment & Rearrange", it always outperformed other three control groups for most simulation instances, except for PD-50. Observe that the values of HV obtained by ALNS+NSGA-II under "Segment & Rearrange" were always the largest for all simulation instances. Thus, "Segment & Rearrange" clearly outperformed and turned out to be the winner.

---

[5] To highlight the value gap between their HV significantly, we multiply the value of HV by 1000.

Table 4 The compared results under four different groups

| Instances (NO. M) | Segment & Rearrange | | | Segment & FOFD | | | Unsegment & Rearrange | | | Unsegment & FOFD | | |
|---|---|---|---|---|---|---|---|---|---|---|---|---|
| | $HV$ | $\overline{v_1}$ | $\overline{v_2}$ | $HV$ | $\overline{v_1}$ | $\overline{v_2}$ | $HV$ | $\overline{v_1}$ | $\overline{v_2}$ | $HV$ | $\overline{v_1}$ | $\overline{v_2}$ |
| ND-50 | 316.84 | 0.167 | 0.6689 | 207.54 | 0.3163 | 0.724 | 309.49 | 0.1799 | 0.7096 | 207.27 | 0.3239 | 0.7121 |
| ND-100 | 472.81 | 0.1541 | 0.566 | 196.71 | 0.4424 | 0.777 | 458.27 | 0.1653 | 0.5871 | 191.17 | 0.4383 | 0.7713 |
| ND-150 | 569.64 | 0.2922 | 0.372 | 160.73 | 0.6045 | 0.7297 | 530.49 | 0.2967 | 0.4088 | 159.42 | 0.613 | 0.7176 |
| ND-200 | 629.68 | 0.2648 | 0.25 | 158.65 | 0.5966 | 0.7188 | 596.52 | 0.3109 | 0.2513 | 152.99 | 0.6281 | 0.6933 |
| ND-250 | 551.76 | 0.3672 | 0.2631 | 133.43 | 0.6774 | 0.6897 | 496.99 | 0.3848 | 0.2934 | 128.1 | 0.6946 | 0.6794 |
| ND-300 | 493.92 | 0.4434 | 0.2462 | 110.45 | 0.7231 | 0.6902 | 470.88 | 0.468 | 0.2424 | 112.69 | 0.7247 | 0.6945 |
| ND-350 | 437.18 | 0.5126 | 0.2445 | 89.65 | 0.7885 | 0.6841 | 399.07 | 0.5526 | 0.2539 | 90.09 | 0.7834 | 0.6982 |
| ND-400 | 421.3 | 0.5365 | 0.2419 | 89.45 | 0.7948 | 0.6926 | 403.42 | 0.5535 | 0.2514 | 90.68 | 0.7744 | 0.7067 |
| ND-450 | 395.82 | 0.5858 | 0.2311 | 83.83 | 0.7971 | 0.6854 | 368.38 | 0.5936 | 0.2544 | 86.56 | 0.782 | 0.7239 |
| ND-500 | 362.96 | 0.5962 | 0.2215 | 69.98 | 0.8274 | 0.7093 | 355.42 | 0.5972 | 0.232 | 66.06 | 0.8334 | 0.7066 |
| PD-50 | 372.43 | 0.025 | 0.7184 | 225.81 | 0.2722 | 0.7549 | 374.48 | 0.025 | 0.717 | 226.24 | 0.263 | 0.7592 |
| PD-100 | 502.76 | 0.1946 | 0.5164 | 191.99 | 0.4952 | 0.7215 | 485.57 | 0.1989 | 0.53 | 191.25 | 0.4977 | 0.7137 |
| PD-150 | 559.96 | 0.239 | 0.4253 | 166.67 | 0.6019 | 0.6905 | 532.71 | 0.2787 | 0.4599 | 174.18 | 0.5627 | 0.7179 |
| PD-200 | 498.98 | 0.3876 | 0.3842 | 134.88 | 0.6807 | 0.6881 | 469.78 | 0.3894 | 0.412 | 142.45 | 0.6821 | 0.6626 |
| PD-250 | 488.29 | 0.4289 | 0.3263 | 118.11 | 0.7341 | 0.6789 | 467.34 | 0.4368 | 0.3263 | 116.92 | 0.7335 | 0.6789 |

| Instances | Segment & Rearrange | | | Segment & FOFD | | | Unsegment & Rearrange | | | Unsegment & FOFD | | |
|---|---|---|---|---|---|---|---|---|---|---|---|---|
| (NO. M) | $HV$ | $\overline{v_1}$ | $\overline{v_2}$ | $HV$ | $\overline{v_1}$ | $\overline{v_2}$ | $HV$ | $\overline{v_1}$ | $\overline{v_2}$ | $HV$ | $\overline{v_1}$ | $\overline{v_2}$ |
| PD-300 | 431.23 | 0.4938 | 0.3121 | 87.15 | 0.7941 | 0.6724 | 420.96 | 0.5062 | 0.332 | 90.34 | 0.7986 | 0.6609 |
| PD-350 | 436.85 | 0.4957 | 0.2885 | 83.75 | 0.7976 | 0.7097 | 407.76 | 0.5103 | 0.2934 | 82.33 | 0.7923 | 0.7039 |
| PD-400 | 421.58 | 0.5139 | 0.2756 | 78.6 | 0.8215 | 0.7063 | 416.54 | 0.5287 | 0.311 | 77.34 | 0.8289 | 0.6921 |
| PD-450 | 372.76 | 0.5758 | 0.2873 | 65.49 | 0.8522 | 0.6837 | 363.81 | 0.576 | 0.3157 | 69.57 | 0.8398 | 0.6797 |
| PD-500 | 320.05 | 0.6378 | 0.2916 | 61.79 | 0.8562 | 0.6937 | 311.96 | 0.6463 | 0.3139 | 60.98 | 0.8683 | 0.6439 |
| MD-100 | 412.62 | 0.0917 | 0.6448 | 207.45 | 0.3457 | 0.7749 | 409.15 | 0.0959 | 0.6554 | 205.98 | 0.3565 | 0.7772 |
| MD-200 | 563.55 | 0.1681 | 0.4526 | 178.13 | 0.5136 | 0.7549 | 547.41 | 0.1837 | 0.474 | 185.03 | 0.5144 | 0.7385 |
| MD-300 | 513.92 | 0.3465 | 0.3584 | 131.72 | 0.6452 | 0.7507 | 496.03 | 0.3474 | 0.4394 | 133.06 | 0.6352 | 0.7415 |
| MD-400 | 463.54 | 0.4621 | 0.3176 | 108.03 | 0.7309 | 0.7165 | 439.27 | 0.4683 | 0.3738 | 104.44 | 0.7245 | 0.7228 |
| MD-500 | 397.26 | 0.5433 | 0.3139 | 85.34 | 0.7774 | 0.7422 | 380.26 | 0.5466 | 0.3673 | 86.39 | 0.7754 | 0.7262 |
| MD-600 | 379.19 | 0.5871 | 0.3128 | 73.98 | 0.795 | 0.7594 | 353.16 | 0.5917 | 0.3686 | 74.64 | 0.8026 | 0.7483 |
| MD-700 | 328.98 | 0.6186 | 0.3246 | 65 | 0.8271 | 0.7073 | 318.61 | 0.628 | 0.3737 | 65.51 | 0.8304 | 0.6981 |
| MD-800 | 296.5 | 0.6627 | 0.3062 | 53.57 | 0.8578 | 0.7329 | 283.54 | 0.67 | 0.3529 | 55.34 | 0.8448 | 0.7363 |
| MD-900 | 275.36 | 0.6846 | 0.3196 | 47.08 | 0.8711 | 0.7514 | 262.32 | 0.696 | 0.3577 | 50.58 | 0.8622 | 0.7385 |
| MD-1000 | 246.18 | 0.7086 | 0.3266 | 45.08 | 0.8754 | 0.7608 | 238.42 | 0.7266 | 0.3834 | 47.51 | 0.8678 | 0.7415 |

To show the efficiency of ALNS+NSGA-II under the four groups visually, four simulation instances are chosen. The Pareto frontier and the iteration values of HV obtained by ALNS+NSGA-II under the four groups after 200 iterations are presented in Figure 4. Note that, the black dotted line, the blue dotted line, the red dotted line and the green dotted line correspond to "Segment & Rearrange", "Segment & FOFD", "Unsegment & Rearrange" and "Unsegment & FOFD" respectively.

From the figures it is clear that the efficiency of ALNS+NSGA-II under "Rearrange" is significantly better than that of ALNS+NSGA-II under "FOFD", since the location of the Pareto frontiers and the iteration values of HV obtained by ALNS+NSGA-II under "Rearrange" are significantly better than those under "FOFD".

The efficiency of ALNS+NSGA-II under "Segment" is similar to that of ALNS+NSGA-II under "Unsegment". But "Segment & Rearrange" is slightly better than "Segment & FOFD" for different scales of simulation instance, especially with the scale of simulation instance increases, this difference becomes more prominent.

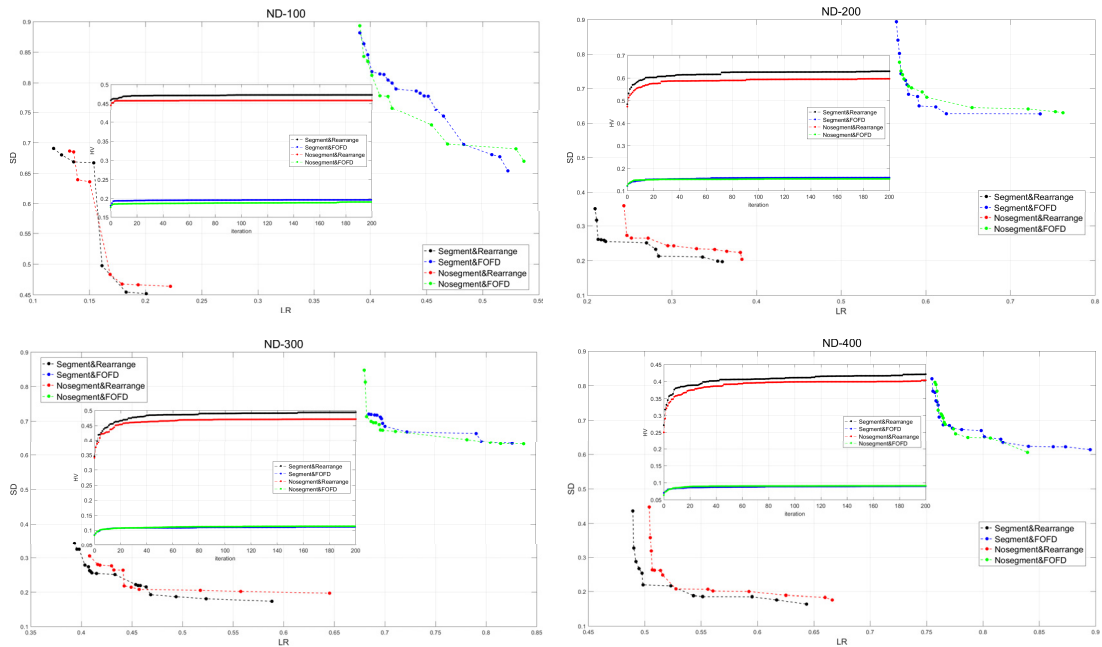

Figure 4 The Pareto frontier and the iteration values of HV obtained by ALNS+NSGA-II under the four groups

In addition, according to the trend of change in of the values of iteration HV, we also find the convergence of ALNS+NSGA-II was always consistent, regardless of the experiment groups. The value of HV obtained by ALNS+NSGA-II always reaches a stable value after about 10 iterations for the four simulation instances.

All in all, "Segment & Rearranger" is not only good for transmitting more original image data, it is also effective in using transmission windows fully and evenly. Therefore, we restricted our attention only to "Segment & Rearrange" for further experiments.

## 4.3. The evolution of ALNS+NSGA-II

In this section, we analyze the evolution of ALNS+NSGA-II in detail. We first analyze the evolutionary mechanism of ALNS+NSGA-II. Then we analyze the effect of the size of the taboo list $B$ on the efficiency of ALNS+NSGA-II, and the evolution of all proposed operators.

**4.3.1. The evolutionary mechanism**

Since there is evidence showing that a random search can be competitive to evolutionary approaches in multi-objective spaces (Wang et al., 2013), we propose a crude random evolutionary mechanism (CREM), in which the elitist solutions are preserved randomly, to analyze the evolutionary mechanism, NSGA-II, in our algorithm. Then, combine CREM with ALNS (ALNS+CREM) to develop a control algorithm.

Based on the test instances in Normal distribution (ND), we restarted the two algorithms, ALNS+NSGA-II and ALNS+CREM, respectively for 50 times. A boxplot of Hypervolume (HV) obtained by them is shown in Figure 5. The black boxes and blue boxes denote the HV obtained by them respectively, and the red pluses indicate the outliers. In addition, in order to display the efficiency of NSGA-II visually, we also plot the Pareto frontiers with the best value of HV during 50 times restarting for five specific simulation instances, ND-100~ND-500, with an increasement step of 100.

It discloses that the nondominated solution obtained by ALNS+NSGA-II is consistently better than the nondominated solution obtained by ALNS+CSEM as the location of black boxes is always higher than the location of blue boxes for all simulation instances.

Also, the length of black boxes is significantly shorter than that of blue boxes, and the number of red pluses for black boxes is always fewer than that for the blue boxes, both of which reflect that ALNS+NSGA-II can achieve the nondominated solutions in a stable way.

For the five specific test instances, the location of Pareto frontiers also shows that NSGA-

II is a nice evolutionary mechanism. On the one hand, the location of Pareto frontiers obtained by ALNS+NSGA-II is always under that obtained by ALNS+CSEM. Further, the Pareto frontiers obtained by ALNS+NSGA-II is also consistently longer. Note that the longer the Pareto frontier is, the more diverse the solutions are. These two aspects establish that NSGA-II is a good evolutionary mechanism for our problem.

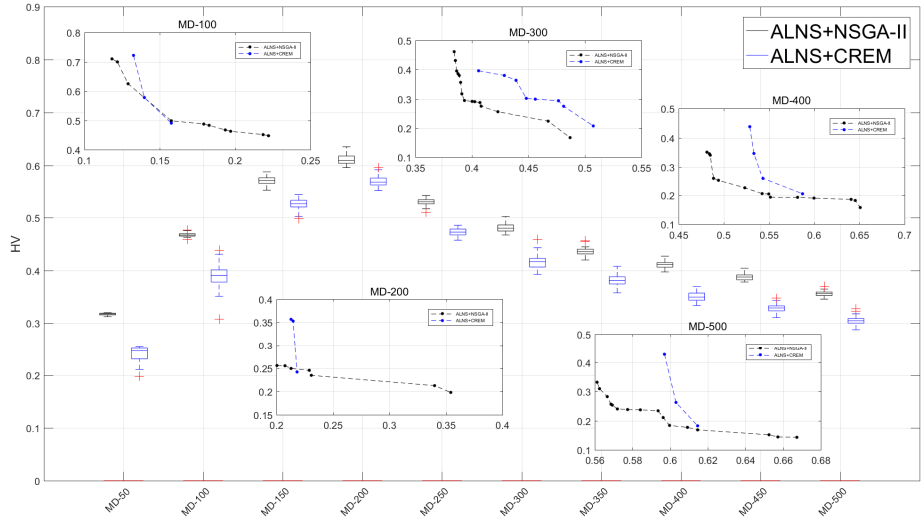

Figure 5 Boxplot: the values of HV obtained by ALNS+NSGA-II and ALNS+CREM, and Dotted line: the best Pareto frontiers of five specific simulation instances obtained by these two algorithms

In addition, as mentioned above, ALNS+NSGA-II also has perfect convergence, which can reach the stable high value of HV within a few iterations. Therefore, considering both, the efficiency and the convergence, NSGA-II is a suitable evolutionary mechanism for D-SIDSP.

**4.3.2. The effect of adaptive size of $B$ on ALNS+NSGA-II**

As mentioned earlier, the taboo list ($B$) is critical for the adaptive large neighborhood search algorithm (ALNS). To analyze the effect of different size of $B$, that is $TR$, on the performance of ALNS+NSGA-II, we carried out two types of experiments, called "Static size" and "Adaptive size", respectively, referring to the nature of the taboo list size. For the "Static size", the value of $TR$ is static and belongs to the interval $[0,1]^6$, with a step is chosen as 0.1. For the "Adaptive size", the value of $TR$ is adaptive and belongs some intervals. Ten adaptive intervals are considered, in which the right values are all equal to 0, and the left values belongs to the interval $[0.1,1]$, with a step-size as 0.1. We considered four different instances from the Polar distribution (PD), PD-50~PD-350, with a step as 100. In addition, we set $MaxIter = 50$ and restart ALNS+NSGA-II 50 times for each instance. The final values of HV and the running time based on all experiments for all test instances are drawn using boxplots in Figure 6. There are two types of boxplots for every instance, one used to represent the values of HV (the blue boxplot) and the other to represent the running time (the black boxplot).

---

[6] The value of $TR$ equals zero means the taboo list is invalid, and the evolution is only depended on the "Repair" operators rank all available original image data in ALNS.

We observed that the values of HV obtained by ALNS+NSGA-II under the "Adaptive size" taboo list are always better than that under "Static size" type for all instances. The running time spent by ALNS+NSGA-II under "Adaptive size" is also significantly less than that under "Static size". Therefore, it appears that a taboo list with an adaptive size is a better choice in obtaining good quality solutions faster.

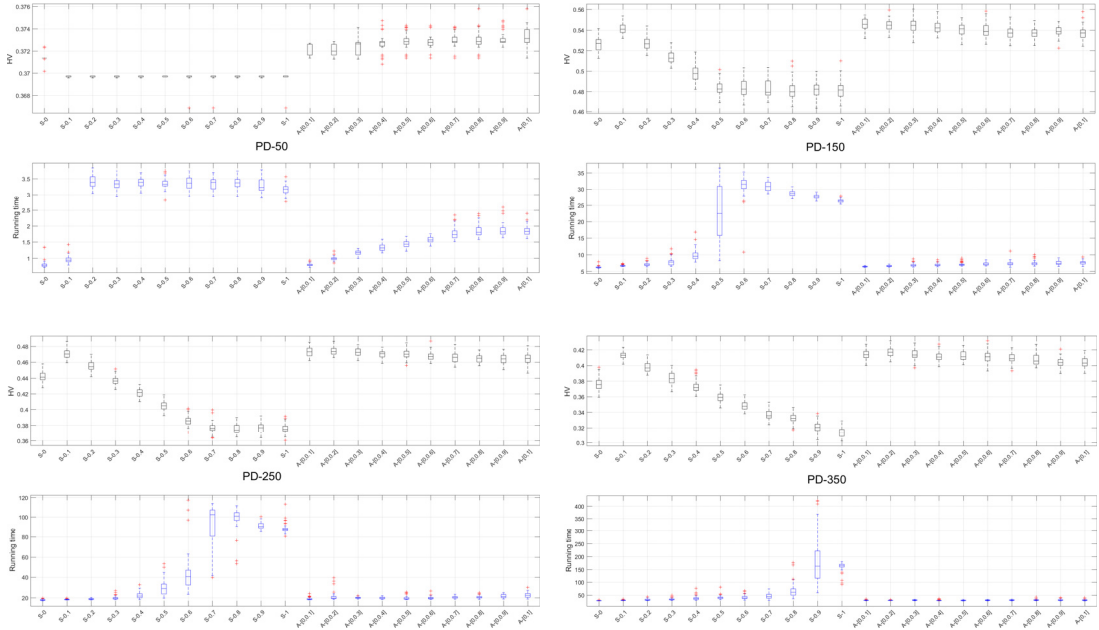

Figure 6 The restarting values of HV obtained by ALNS+NSGA-II under these four simulation instances after 50 times and the running time of them

The flexibility of ALNS+NSGA-II under "Static size" is worse than that under "Adaptive size". For smaller instances, like PD-50 and PD-100, the available transmission windows are sufficient to receiving all original image data, so the values of HV and running time under different size of taboo list are similar. But, as the size of instance increases, the quality of ALNS+NSGA-II under "Static size" deteriorates, resulting in lower HV values and increased running times.

All in all, ALNS+NSGA-II under "Adaptive size" always performed better with higher HV

value and reasonable running times, regardless the size of the instance. Thus, we have decided to use the taboo list with "Adaptive size" belongs to the interval [0,0.2] in ALNS+NSGA-II for all subsequent experiments.

### 4.3.3. The evolution of all operators

A simulation instance, MD-200 in the Mixed distribution, is chosen to analyze the evolution of all our proposed operators: "Destroy" operators and "Repair" operators. Let $\lambda$ belongs to the interval [0,1] with steps 0.1, and restart ALNS+NSGA-II for 50 times based on MD-500. The final weights of all operators under different values of $\lambda$ are drawn by two boxplots in Figure 7 respectively. The y-axis denotes the values of final weight, while the x-axis represents different values of $\lambda$. Note that the scale of the y-axis in the two subplots is different.

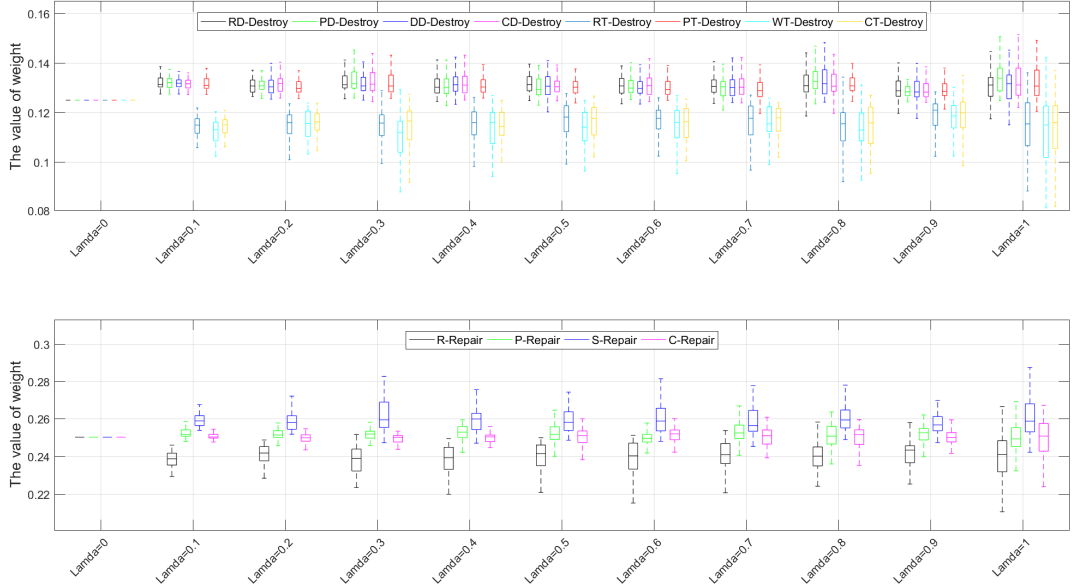

Figure 7 Boxplot: the final weight of all operators ("Destroy" operators and "Repair" operators) after restarting ALNS+NSGA-II for 50 times based on MD-200 under different values of $\lambda$

The average values of the final weight for all operators ("Destroy" operators and "Repair" operators) under different values of $\lambda$ are almost located in the same horizontal line around the average value respectively. For "Destroy" operators, it is 0.125 and for "Repair" operators, it is 0.25. This phenomenon reflects that the efficiency of all proposed operators under different values of $\lambda$ is similar (so we set $\lambda = 0.5$ for our experiments).

In addition, **RD-Destroy**, **PD-Destroy**, **DD-Destroy**, **CD-Destroy** and **PT-Destroy** are slightly better than **RT-Destroy**, **WT-Destroy** and **CT-Destroy** under all of the distinct values of $\lambda$, while **P-Repair** was constantly better among all "Repair" operators under all different values of $\lambda$.

Note that the length of a box represents the deviation. Since the boxes of "Destroy" operators are always longer than that of "Repair" operators, the efficiency of the "Repair" operators is more stable with the values of $\lambda$ changes.

As mentioned earlier, "Destroy" operators and "Repair" operators should be used in pair. When $\lambda$ equals 0.5, the average of final weight of **DD-Destroy** is slightly bigger among "Destroy" operators, while the mean of final weight of **S-Repair** is significantly larger among "Repair" operators, so adopting **P-Destroy** and **P-Repair** in pair is more likely to generate better solutions to be some extent.

## 4.4. The efficiency of ALNS+NSGA-II

In (Karapetyan et al., 2015) a greedy randomized adaptive search algorithm (GRASP), ejection chain algorithm (EC), simulated annealing algorithm (SA) and tabu search algorithm (TS), are considered to solve the traditional SIDSP. The first three of them[7] are based on adaptive large search neighborhoods (ALNS) and dependent on a swap operator to evolve the solutions. Therefore, we modify them, add ***ODCS*** () as defined in Algorithm 2 and adopt NSGA-II as the evolutionary mechanism, and compared the algorithms to analyze the efficiency of ALNS+NSGA-II. In addition, all simulation instances in Mixed distribution (MD) are considered for this experiment. Considering that the local search method of them are different, the same maximum running time is set for controlling these four algorithms and same initial solutions will also be used for them, for every simulation instance. The Pareto frontier and iteration values of HV obtained by these four algorithms are shown in Figure 8. Ten subplots denote the results based on MD-100~MD-1000 respectively (The bigger one denotes the Pareto frontier, while the small one represents the values of HV).

It follows from the results of our computational experiments that ALNS+NSGA-II clearly outperformed the other three algorithms no matter what the size of the test instance are. Further, as the size of the test instances increases, the difference become more significant. Observe that the Pareto frontier obtained by ALNS+NSGA-II is always located under that obtained by the other three algorithms. On the other hand, the iteration values of HV obtained by ALNS+NSGA-II are always above that of the other three algorithms.

---

[7] Their experiments have shown that such a TS implementation performs poorly on DRPP, so we do not consider TS as a compared algorithm.

GRASP+NSGA-II, EC+NSGA-II and SA+NSGA-II are based on a swap operator to generate offspring solutions (Karapetyan et al., 2015), and the swap operator only changes the transmission scheme of an original image data in one time. In fact, these three algorithms can be viewed as special cases of ALNS+NSGA-II. Thus, it is inevitable that ALNS+NSGA-II is consistently better these three algorithms.

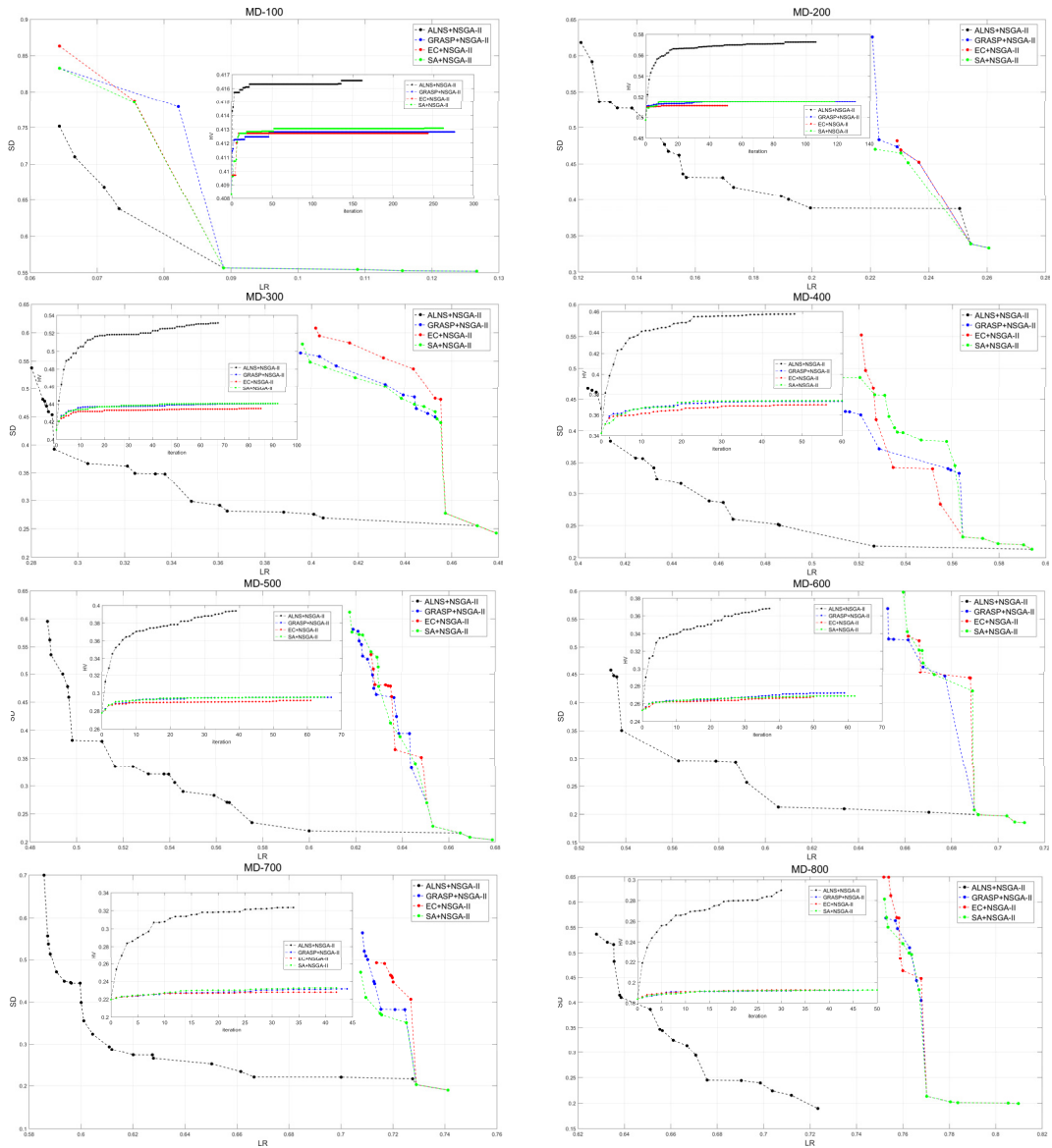

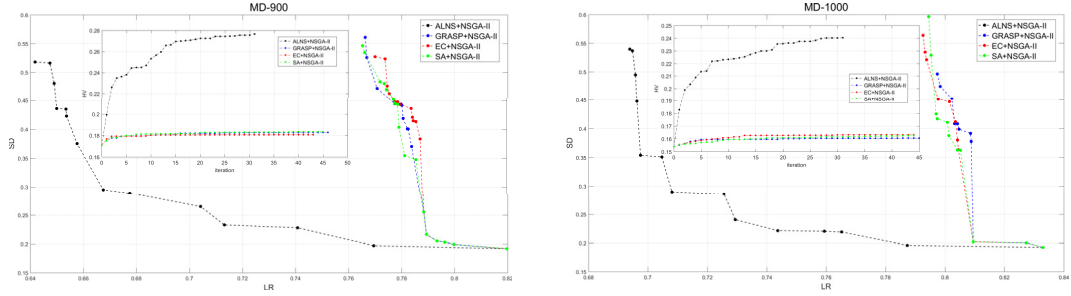
Figure 8 The Pareto frontier and the iteration values of HV obtained by these four algorithms

## 5. Conclusions

In this paper, we formulated a dynamic two-phase satellite image data downlink scheduling problem (D-SIDSP) as a bi-objective optimization problem, optimizing the image data transmission and the service-balance degree, and developed an adaptive bi-objective memetic algorithm, ALNS+NSGA-II. The algorithm uses an adaptive large neighborhood search algorithm (ALNS) as the local search scheme to breed offspring solutions and a nondominated sorting genetic algorithm II (NSGA-II) as the evolutionary mechanism to achieve Pareto frontier faster. In addition, a random greedy heuristic algorithm (RGHA) is designed as the initialization algorithm, and two types operators, "Destroy" and "Repair", with an adaptive size taboo bank are used to generate adaptive large neighborhoods (ALNs). In order to analyze the influence of "Segment & Rearrange" on D-SIDSP and the evolution and efficiency of ALNS+NSGA-II, different classes of test instances are generated based on real world situations and consistent with current research practices in the field.

Comparing with the three candidate control groups, "Segment & Rearrange" exhibited superior performance based on various performance metrics.

In addition, several standard evolutionary algorithms, including GRASP+NSGA-II, EC+NSGA-II and SA+NSGA-II, were compared with our ALNS+NSGA-II. The experiments disclosed the superiority of our algorithm ALNS+NSGA-II and hence it is recommended as a viable approach to solve D-SIDSP. In addition to the algorithmic advancements, the novelty of our work also includes the introduction of D-SIDSP as opposed to traditional SIDSP. The D-SIDSP model offers additional flexibility in handling complex downlink scheduling problem for modern EOS clusters.

**Acknowledgements:** The research of Zhongxiang Chang was supported by the science and

# References

BARBULESCU, L., WATSON, J.-P., WHITLEY, L. D. & HOWE, A. E. 2004. Scheduling Space-Ground Communications for the Air Force Satellite Control Network. *Journal of Scheduling,* 1**,** 7-34.

BRADSTREET, L., WHILE, L. & BARONE, L. 2008. A Fast Incremental Hypervolume Algorithm. *IEEE Transactions on Evolutionary Computation,* 12**,** 714-723.

CHANG, Z., CHEN, Y., YANG, W. & ZHOU, Z. Analysis of Mission Planning Problem for Video Satellite Imaging with Variable Imaging Duration.  2019 IEEE Symposium Series on Computational Intelligence, SSCI 2019, 2019 Xiamen, China. IEEE, 1700-1707.

CHANG, Z., CHEN, Y., YANG, W. & ZHOU, Z. 2020. Mission planning problem for optical video satellite imaging with variable image duration: a greedy algorithm based on heuristic knowledge. *Advances in Space Research,* 66**,** 2597-2609.

CHANG, Z., ZHOU, Z., XING, L. & YAO, F. 2021a. Integrated scheduling problem for earth observation satellites based on three modeling frameworks: an adaptive bi-objective memetic algorithm. *Memetic Computing,* 13**,** 203-226.

CHANG, Z., ZHOU, Z., YAO, F. & LIU, X. 2021b. Observation scheduling problem for AEOS with a comprehensive task clustering. *Journal of Systems Engineering and Electronics,* 32**,** 347-364.

DEB, K., PRATAP, A., AGARWAL, S. & MEYARIVAN, T. 2002. A fast and elitist multiobjective genetic algorithm: NSGA-II. *IEEE Transactions on Evolutionary Computation,* 6**,** 182-197.

DU, Y., XING, L., ZHANG, J., CHEN, Y. & HE, Y. 2019. MOEA based memetic algorithms for multi-objective satellite range scheduling problem. *Swarm and Evolutionary Computation,* 50.

GONG, M. G., JIAO, L. C., YANG, D. D. & MA, W. P. 2009. Research on evolutionary multi-objective optimization algorithms. *Journal of Software,* 2**,** 271−289.

GUO, H., LIU, J., LI, A. & ZHANG, J. 2012. Earth observation satellite data receiving, processing system and data sharing. *International Journal of Digital Earth,* 5**,** 241-250.

HAMACHER, H. W., PEDERSEN, C. R. & RUZIKA, S. 2007. Finding representative systems for discrete bicriterion optimization problems. *Operations Research Letters,* 35**,** 336-344.

HE, L., DE WEERDT, M. & YORKE-SMITH, N. 2019. Time/sequence-dependent scheduling: the design and evaluation of a general purpose tabu-based adaptive large neighbourhood search algorithm. *Journal of Intelligent Manufacturing,* 31**,** 1051-1078.

HE, L., LIU, X. L., LAPORTE, G., CHEN, Y. W. & CHEN, Y. G. 2018. An improved adaptive large neighborhood search algorithm for multiple agile satellites scheduling. *Computers & Operations Research,* 100**,** 12-25.

HUANG, W., SUN, S. R., JIANG, H. B., GAO, C. & ZONG, X. Y. 2018. GF-2 Satellite 1m/4m Camera Design and In-Orbit Commissioning. *Chinese Journal of Electronics,* 27**,** 1316-1321.

JAKHU, R. S. & PELTON, J. N. 2014. The Development of Small Satellite Systems and Technologies. *Small Satellites and Their Regulation.*

JAWAK, S. D. & LUIS, A. J. 2013. Improved land cover mapping using high resolution multiangle 8-band WorldView-2 satellite remote sensing data. *Journal of Applied Remote Sensing,* 7.

JONES, A. 2018. First Chinese launch of 2018 puts two SuperView-1 observation satellites into low Earth orbit.

KADZIŃSKI, M., TERVONEN, T., TOMCZYK, M. K. & DEKKER, R. 2017. Evaluation of multi-objective optimization approaches for solving green supply chain design problems. *Omega,* 68**,** 168-184.

KARAPETYAN, D., MITROVIC MINIC, S., MALLADI, K. T. & PUNNEN, A. P. 2015. Satellite downlink scheduling problem: A case study. *Omega,* 53**,** 115-123.

KIDD, M. P., LUSBY, R. & LARSEN, J. 2020. Equidistant representations: Connecting coverage and uniformity in discrete biobjective optimization. *Computers & Operations Research,* 117.

LI, J., LI, J., CHEN, H. & JING, N. 2014. A data transmission scheduling algorithm for rapid-response earth-observing operations. *Chinese Journal of Aeronautics,* 27**,** 349-364.

LIU, X. L., LAPORTE, G., CHEN, Y. W. & HE, R. J. 2017. An adaptive large neighborhood search metaheuristic for agile satellite scheduling with time-dependent transition time. *Computers & Operations Research,* 86**,** 41-53.

LU, S., CHANG, Z., ZHOU, Z. & YAO, F. An Adaptive Multi-objective Memetic Algorithm: a Case of Observation Scheduling for Active-imaging AEOS.  2021 7th International Conference on Big Data and Information Analytics (BigDIA), 2021. 285-294.

LUO, K., WANG, H., LI, Y. & LI, Q. 2017. High-performance technique for satellite range scheduling. *Computers & Operations Research,* 85**,** 12-21.

MALLADI, K. T., MINIC, S. M., KARAPETYAN, D. & PUNNEN, A. P. 2016. Satellite Constellation Image Acquisition Problem: A Case Study. *In:* FASANO, G. & PINTéR, J. D. (eds.) *Space Engineering. Springer Optimization and Its Applications.* Cham: Springer.

MARINELLI, F., NOCELLA, S., ROSSI, F. & SMRIGLIO, S. 2011. A Lagrangian heuristic for satellite range scheduling with resource constraints. *Computers & Operations Research,* 38**,** 1572-1583.

MUTER, İ. & SEZER, Z. 2018. Algorithms for the one-dimensional two-stage cutting stock problem. *European Journal of Operational Research,* 271**,** 20-32.

NA, C. 2016. China's First Overseas Land Satellite Receiving Station Put into Operation.

NERI, F. & COTTA, C. 2012. Memetic algorithms and memetic computing optimization: A literature review. *Swarm and Evolutionary Computation,* 2**,** 1-14.

PEI-JUN, A., XUE-MEI, W., ZHI-QING, Z. & FENG, G. 2008. Distribution of Overseas Satellites Ground Stations and Their Operational Characteristics. *Remote Sensing Technology and Application,* 7**,** 23-43.

ROPKE, D. P. S. 2007. A general heuristic for vehicle routing problems. *Computers & Operations Research,* 34**,** 2403-2435.

SAI, W., REN, J. & JIDONG, Z. 2018. Super View-1-China's First Commercial Remote Sensing Satellite Constellation with a High Resolution of 0.5 m. *Aerospace China,* 1**,** 30-38.

SHE, Y., LI, S., LI, Y., ZHANG, L. & WANG, S. 2019. Slew path planning of agile-satellite antenna pointing mechanism with optimal real-time data transmission performance. *Aerospace Science and Technology,* 90**,** 103-114.

SRINIVAS, N. & DEB, K. 1995. Multi-objective function optimization using nondominated sorting genetic algorithms. *Evolutionary Computation,* 3**,** 221-248.

VAZQUEZ, A. J. & ERWIN, R. S. 2014. On the tractability of satellite range scheduling. *Optimization Letters,* 9**,** 311-327.

WANG, P. & REINELT, G. 2010. A Heuristic for an Earth Observing Satellite Constellation Scheduling Problem with Download Considerations. *Electronic Notes in Discrete Mathematics,* 36**,** 711-718.

WANG, P., REINELT, G., GAO, P. & TAN, Y. 2011. A model, a heuristic and a decision support system to solve the scheduling problem of an earth observing satellite constellation. *Computers & Industrial Engineering,* 61**,** 322-335.

WANG, R., PURSHOUSE, R. C. & FLEMING, P. J. 2013. Preference-Inspired Coevolutionary Algorithms for Many-Objective Optimization. *IEEE Transactions on Evolutionary Computation,* 17**,** 474-494.

WANG, X., WU, G., XING, L. & PEDRYCZ, W. 2020. Agile Earth observation satellite scheduling over 20 years: formulations, methods and future directions. *IEEE Systems Journal*.

WU, X. & CHE, A. 2019. A memetic differential evolution algorithm for energy-efficient parallel machine scheduling. *Omega,* 82**,** 155-165.

WU, X. & CHE, A. 2020. Energy-efficient no-wait permutation flow shop scheduling by adaptive multi-objective variable neighborhood search. *Omega,* 94.

XIAO, Y., ZHANG, S., YANG, P., YOU, M. & HUANG, J. 2019. A two-stage flow-shop scheme for the multi-satellite observation and data-downlink scheduling problem considering weather uncertainties. *Reliability Engineering & System Safety,* 188**,** 263-275.

ZHANG, J., XING, L., PENG, G., YAO, F. & CHEN, C. 2019. A large-scale multiobjective satellite data transmission scheduling algorithm based on SVM+NSGA-II. *Swarm and Evolutionary Computation,* 50.

ZUFFEREY, N., AMSTUTZ, P. & GIACCARI, P. 2008. Graph colouring approaches for a satellite range scheduling problem. *Journal of Scheduling,* 11**,** 263-277.